\documentclass[10pt, a4paper]{article}

\usepackage{fontspec}

\usepackage{lrec-coling2024} 

\usepackage{booktabs} 
\usepackage{subcaption} 
\usepackage{float}    
\usepackage{placeins} 
\usepackage{array}     
\usepackage{graphicx}  
\newcolumntype{P}[1]{>{\raggedright\arraybackslash}p{#1}}
\usepackage{latexsym}
\usepackage{amsmath}
\usepackage{todonotes}
\usepackage{xcolor}
\usepackage{stfloats}
\usepackage{newtxmath}  
\usepackage[utf8]{inputenc}

\title{A Typologically Grounded Evaluation Framework for Word Order and Morphology Sensitivity in Multilingual Masked LMs}

\name{Anna Feldman, Libby Barak, Jing Peng} 

\address{Montclair State University \\
         New Jersey, USA \\
         \{feldmana, barakl, pengj\}@montclair.edu\\}

\abstract{
We introduce a typology-aware diagnostic for multilingual masked language models that tests reliance on word order versus inflectional form.
Using Universal Dependencies, we apply inference-time perturbations: full token scrambling, content-word scrambling with function words fixed, dependency-based head--dependent swaps, and sentence-level lemma substitution (+L), which lemmatizes both the context and the masked target label.
We evaluate mBERT and XLM-R on English, Chinese, German, Spanish, and Russian.
Full scrambling drives word-level reconstruction accuracy near zero in all languages; partial and head--dependent perturbations cause smaller but still large drops.
+L has little effect in Chinese but substantially lowers accuracy in German/Spanish/Russian, and it does not mitigate the impact of scrambling.
Top-5 word accuracy shows the same pattern: under full scrambling, the gold word rarely appears among the five highest-ranked reconstructions.
We release code, sampling scripts, and balanced evaluation subsets; Turkish results under strict reconstruction are reported in the appendix.
\\ \newline \Keywords{multilingual evaluation, masked language models, morphology}}

\begin{document}
\maketitleabstract

\section{Introduction}

Multilingual transformer models such as mBERT and XLM-R \citep{conneau-etal-2020-unsupervised} achieve strong cross-linguistic performance, but what kinds of linguistic cues actually drive their predictions remains unclear. Languages differ sharply in how they encode grammatical relations: fixed-order languages rely on strict sequencing (e.g., English, Chinese), while case-marking or richly inflected languages allow greater flexibility through morphology (e.g., Russian, German, Spanish). If multilingual models depend mainly on linear order, they risk an English-centered bias that disadvantages morphologically rich or free-order languages. If they exploit morphology, they should be more resilient under word-order disruption. Understanding this balance is essential for evaluating the typological generalization and fairness of multilingual encoders.

We treat word order and morphology as partially redundant structural cues and ask how strongly current multilingual encoders rely on each under controlled disruption. The goal is a \emph{diagnostic}: to quantify \emph{relative sensitivity} and \emph{non-additivity} across languages, not to crown a winner between order and morphology.

We do not train new models or argue that reliance on multiple cues is inherently bad. Natural languages routinely exhibit redundancy and multiple exponence. Our contribution is an evaluation protocol and typologically grounded stress tests that expose where models are brittle and where cues overlap.

Prior work shows that transformer-based language models exhibit strong positional biases \citep{futrell2019neural,warstadt-etal-2020-blimp-benchmark,papadimitriou-etal-2022-classifying-grammatical}, with scrambling often causing sharp performance drops \citep{ettinger-2020-bert,fung2024wordorder}. However, most analyses are monolingual and focus on English, leaving open whether multilingual models compensate for disrupted order by drawing on morphology.

We present a typologically grounded evaluation framework for probing the relative roles of word order and morphology in multilingual masked language models. The framework uses Universal Dependencies (UD) treebanks to apply controlled perturbations at inference time: (i) word-order permutations and (ii) sentence-level lemma substitution (+L). We test mBERT and XLM-R on five typologically diverse languages (English, Chinese, German, Spanish, Russian), combining +L with multiple permutation levels. To ensure comparability, masking is applied before any transformation so the same target token is evaluated across all conditions.

Full scrambling collapses accuracy in every language, while partial scrambling yields large but language-dependent drops. Lemma substitution (+L) has strongly language-dependent effects: it is nearly identity in Chinese but substantially lowers accuracy in German/Spanish/Russian. It never compensates for lost order; tiny gains in Chinese are consistent with tokenization/reconstruction quirks rather than genuine morphological leverage. Interaction analyses show that the combined effects are typically sub-additive (joint harm is smaller than additive expectations) indicating overlapping rather than complementary cues. Current multilingual encoders are strongly sensitive to word-order disruption in this diagnostic setting; lemma-normalized contexts (+L) do not yield robustness when order cues are degraded.
Because +L is a coarse, sentence-level normalization that changes the label space (surface form vs.\ lemma), we interpret it as an inflection-stripping diagnostic rather than a feature-specific test of case/agreement/tense.


\section{Related Work} \label{sec:related_work}

Neural language models represent syntactic structure differently depending on architecture and training data. While RNNs struggle with long-range dependencies \citep{linzen2016assessing,gulordava-etal-2018-colorless}, transformers leverage self-attention mechanisms that, in principle, allow them to capture hierarchical syntax \citep{vaswani2017attention}. In practice, however, transformers exhibit strong positional biases, often relying heavily on surface-level word order \citep{futrell2019neural,warstadt-etal-2020-blimp-benchmark,papadimitriou-etal-2022-classifying-grammatical}.

One line of work probes these biases by scrambling word order either during training or inference. Some studies show that models pretrained on shuffled text can still achieve strong performance, suggesting a surprising degree of robustness to word order disruption \citep{sinha-etal-2021-masked,hessel-schofield-2021-effective,gupta-jaggi-2021-obtaining}. However, other work finds that performance is highly sensitive to the type of perturbation and the level of linguistic abstraction required by the task \citep{chen2024doeswordordermatter,papadimitriou-etal-2022-classifying-grammatical}. For example, inference-time scrambling often leads to sharp performance drops in tasks requiring syntax-aware reasoning \citep{pham-etal-2019,oconnor-andreas-2021-context}, though most such work is limited to English or monolingual settings.

Multilingual analyses of word order sensitivity remain rare. Some work investigates scrambling in translation or syntactic transfer \citep{ahmad2019context,liu2020mixed}, while others use token permutation to test cross-linguistic generalization \citep{zhao-etal-2020-limitations}. \citet{yang-etal-2019-convolutional} found that attention-based models are fragile under token swaps, and \citet{ettinger-2020-bert} showed that scrambling impairs masked token prediction even without downstream fine-tuning. Attempts to mitigate this brittleness include introducing de-scrambling objectives \citep{wang2020deshuffle} or auxiliary training on reordered inputs.

Prior probes mostly use local swaps or constrained shuffles \citep{papadimitriou-etal-2022-classifying-grammatical,fung2024wordorder}, with some exploring stronger/global permutations \citep{chen2024doeswordordermatter}; dependency-aware scrambling remains rare.
We add a dependency-aware \textit{Head} condition that swaps a UD head with one of its dependents, targeting predicate-argument anchors while keeping other tokens fixed. We use gold UD trees for cross-lingual comparability, though this can still reduce well-formedness. 

Our work extends this line of inquiry in four key ways: 
\begin{enumerate} \itemsep-4pt
    \item We test word order sensitivity across five typologically diverse languages with varying morphological richness and syntactic flexibility (see Table~\ref{tab:balancedNs} for per-language balanced $N$).
    \item We evaluate two widely used multilingual masked language models, mBERT and XLM-R, under identical perturbation settings (results summarized in Tables~\ref{tab:acc_mb}-\ref{tab:acc_xlm}).
   \item We include a coarse inflection-stripping diagnostic via a sentence-level lemma substitution (+L) that replaces all tokens with their UD \texttt{LEMMA} and evaluates the lemma of the masked target; this probes behavior in lemmatized contexts (condition examples in Table~\ref{tab:cond_examples_en}; corresponding columns in Tables~\ref{tab:acc_mb}-\ref{tab:acc_xlm}).

    \item We probe graded order disruption with three manipulations: (i) \textit{Full} (all UD word tokens permuted), (ii) \textit{Part} (content words permuted; function words fixed), and (iii) \textit{Head} (each UD head swapped with one of its dependents).  (Definitions/examples are in Table~\ref{tab:cond_examples_en}; results are in Tables~\ref{tab:acc_mb}-\ref{tab:acc_xlm})

\end{enumerate}

By combining structured word order manipulations with lemma normalization across diverse languages, we offer new evidence about how multilingual transformers balance surface structure and grammatical cues in syntactic generalization.

\section{Evaluation Framework}\label{sec:methodology}

We evaluate how word order and surface morphology affect masked language model (MLM) performance using inference-time perturbations applied to two multilingual encoders (mBERT, XLM-R) across five languages: English, German, Spanish, Russian,  and Chinese. Using multilingual encoders keeps architecture and training constant while varying language. The framework and all scripts are released for reproducibility and extension to new languages and models.

\paragraph{Sentence-level lemma substitution (+L).}
“+L” replaces every token with its UD \texttt{LEMMA} (from the CoNLL-U treebank, with identity fallback) and evaluates the lemma of the masked target. Only gold lemmas are used; no external lemmatizer is applied.

\paragraph{Word-order perturbations.}
Three complementary perturbations probe different aspects of word-order sensitivity:
(i) maximal disruption of sequence information,
(ii) disruption that keeps local function-word scaffolding, and
(iii) dependency-aware disruption targeting predicate–argument anchors.
All permutations are deterministic given a (seed, sentence ID) pair; within each seed, the same perturbation is reused across models and conditions.

\begin{itemize}\itemsep-4pt
  \item \textbf{Full}: randomly permutes all UD word tokens, serving as a stress test of positional reliance.
  \item \textbf{Part}: permutes content words (NOUN, PROPN, VERB, ADJ, ADV) while keeping function words fixed, testing whether local scaffolding can partially preserve predictions.

  \item \textbf{Head}: iterate heads in left-to-right surface order; for each head (skipping heads with UPOS=PUNCT), select one eligible dependent (excluding UPOS=PUNCT), using a deterministic RNG keyed by (seed, sentence ID, head index), and swap the surface positions of the head and the selected dependent in the token sequence. Swaps are applied sequentially to a working copy of the token list; the UD tree is not updated.

\end{itemize}
Function words are identified by Universal POS (UPOS) tags as the complement of \{NOUN, PROPN, VERB, ADJ, ADV\}. Punctuation (UPOS = PUNCT) is never permuted. A small language-specific stoplist (for example, articles or auxiliaries) serves only as a safety filter and does not override UPOS definitions.

\paragraph{Evaluated conditions.}
We evaluate Orig, Full, Part, and Head, plus +L counterparts for Orig, Full, and Part. Definitions and examples appear in Table~\ref{tab:cond_examples_en}.

\begin{table}[t]
\centering
\footnotesize
\setlength{\tabcolsep}{6pt}
\renewcommand{\arraystretch}{1.05}
\resizebox{\columnwidth}{!}{%
\begin{tabular}{lll}
\toprule
Condition & Input string (EN) & Gold \\
\midrule
Orig & the[1] scientist[2] \texttt{[MASK]} the[3] books[4] yesterday[5]. & analyzed \\
Orig+L & the[1] scientist[2] \texttt{[MASK]} the[3] book[4] yesterday[5]. & analyze \\
Full & yesterday[5] the[3] the[1] scientist[2] books[4] \texttt{[MASK]}. & analyzed \\
Full+L & yesterday[5] the[3] the[1] scientist[2] book[4] \texttt{[MASK]}. & analyze \\
Part & the[1] books[4] scientist[2] the[3] yesterday[5] \texttt{[MASK]}. & analyzed \\
Part+L & the[1] book[4] scientist[2] the[3] yesterday[5] \texttt{[MASK]}. & analyze \\
Head & the[1] scientist[2] books[4] the[3] \texttt{[MASK]} yesterday[5]. & analyzed \\
\bottomrule
\end{tabular}}
\caption{Illustrative English examples for each condition. Bracketed indices [i] mark token identity for expository purposes only. “+L” applies sentence-level UD lemma substitution; the gold label is the lemma of the masked target.}
\label{tab:cond_examples_en}
\end{table}

\paragraph{Masking protocol.}
Masking is applied before perturbation. For each sentence, one content word is selected and all of its subword pieces are replaced by the model’s mask token (\texttt{[MASK]} for mBERT, \texttt{<mask>} for XLM-R). This guarantees that the same target token is evaluated across conditions; under scrambling the mask moves with the permutation. Scoring is done at the word level by reconstructing the full span from subword predictions. Punctuation (UPOS = PUNCT) and closed-class stoplist items are never selected as targets.

\paragraph{Randomization and runs.}
We run three seeds (1, 2, 3) and average metrics across seeds. For each seed, we deterministically seed (i) target selection and (ii) each perturbation using a function of (seed, sentence ID), so that within a seed the same masked targets and permutations are reused across models and conditions (preventing sampling drift in cross-condition/model comparisons). Across seeds, targets and/or permutations can differ, providing a robustness check against sampling variance. Python, NumPy, and PyTorch are seeded on CPU and GPU for each run.

\paragraph{Scoring.}
mBERT uses WordPiece tokenization and XLM-R uses SentencePiece/BPE. Predicted words are reconstructed from subword outputs using model-specific detokenization rules. A prediction is correct only if all subword pieces match the gold target exactly after Unicode NFKC normalization, case folding for alphabetic scripts, and punctuation removal.

\paragraph{Reconstruction span cap.}
We cap word reconstruction at six subword pieces (max\_span\_pieces = 6) to prevent degenerate targets. This has negligible impact on English, German, Spanish, Russian, or Chinese (balanced $N$ per condition $\approx$ 300). 

\paragraph{Ranking-based evaluation (top-$k$).}
Exact word reconstruction (top-1) can understate partial knowledge when perturbations increase uncertainty without fully removing evidence.
Therefore, in addition to word-level top-1 accuracy, we report word-level top-5 accuracy (\texttt{word\_at\_5}) from the same stored top-k candidate lists in our released JSONL outputs: an item is counted correct at top-5 if the gold word appears anywhere in the model's top-5 reconstructed candidates for the masked span. For multi-piece targets, the top-5 list is defined over complete reconstructed \emph{words} (full-span reconstructions produced by the same detokenization routine as top-1), not over individual subword pieces. This complements top-1 by capturing cases where the gold target remains highly ranked even when it is not the single most probable prediction.

\paragraph{Data sampling.}
We use Universal Dependencies (UD) treebanks \cite{nivre2016universal} for all five languages. Sampling is controlled by fixed seeds, and the same sentence IDs are used across models. We report both unbalanced sets (all valid items) and balanced sets (downsampled to the smallest per-condition count per language) to enable like-for-like comparisons.

\paragraph{Released artifacts.}
The release includes perturbation scripts, per-language sentence ID lists, balanced and unbalanced sampling metadata, configuration files, and raw JSONL outputs for each run. All code and metadata are licensed for reuse; UD text is not redistributed.

\section{Datasets and Sampling}\label{sec:datasets}

We evaluate on five UD treebanks  spanning diverse typological profiles:
English (EWT), German (GSD), Spanish (AnCora), Russian (SynTagRus), and Chinese (GSD).  Typologically, English and Chinese are predominantly fixed SVO with minimal/isolating morphology; Spanish and German are fusional with richer inflection (German has case-marking and V2/SOV-in-subordinates); Russian is highly case-marked with relatively free word order. 
We do not perform any additional training or fine-tuning; all results are inference-time perturbation tests on UD sentences.


\vspace{-2mm}
\paragraph{Sentence selection.}
We sample up to 400 sentences per language to bound compute and keep per-language uncertainty comparable (preventing larger treebanks from dominating). We keep the same sentence IDs across models and seeds.
For each seed, a single content word is selected and masked per sentence (Sec.~\ref{sec:methodology}); each sentence then
receives every perturbation condition (Original, Original+L, and scrambling variants with/without +L).

For Turkish, the final balanced test set is smaller than 250 sentences because many candidates violate piece-length constraints for word-level masking and strict reconstruction; together with fewer sentences that meet all perturbation constraints across conditions, this reduces the overlap needed for balanced downsampling.

\vspace{-2mm}
\paragraph{Balanced vs. unbalanced reporting.}
Because some conditions can be dropped by filtering (e.g., target span too long, no movement under partial scramble),
the raw per-condition counts may differ. We therefore report two views:
(i) \emph{unbalanced} (all valid items retained), and
(ii) \emph{balanced} (per language, we downsample each condition to the smallest per-condition count).
Unless otherwise noted, aggregate figures are from the balanced view to ensure like-for-like comparisons across conditions.

\vspace{-2mm}
\paragraph{Perturbation magnitude (surface change).}
To contextualize the relative strength of each manipulation, we quantify how much each perturbation changes the surface token sequence.
For +L we report the fraction of non-punctuation tokens whose surface form differs from its lemma (token change rate).
For scrambling conditions we report the fraction of non-punctuation tokens that change absolute position relative to the original sentence (position change rate).
These magnitude differences emphasize that +L and scrambling are not matched manipulations; our conclusions, therefore, concern model sensitivity under these specific perturbations rather than equalized information removal.

\begin{table}[t]
\centering
\footnotesize
\setlength{\tabcolsep}{3pt}
\renewcommand{\arraystretch}{1.0}
\resizebox{\columnwidth}{!}{%
\begin{tabular}{lcccc}
\toprule
Lang & +L tok.\ chg & Full pos.\ chg & Part pos.\ chg & Head pos.\ chg \\
\midrule
DE & 0.344 & 0.910 & 0.173 & 0.339 \\
EN & 0.191 & 0.906 & 0.160 & 0.455 \\
ES & 0.350 & 0.945 & 0.195 & 0.452 \\
RU & 0.555 & 0.935 & 0.268 & 0.570 \\
ZH & 0.007 & 0.950 & 0.272 & 0.478 \\
\bottomrule
\end{tabular}}
\caption{Perturbation magnitude on evaluated UD sentences (non-punctuation tokens). +L tok.\ chg: fraction of tokens with FORM $\neq$ LEMMA. Pos.\ chg: fraction of tokens whose absolute position differs from Orig.}
\label{tab:perturb_magnitude}
\end{table}

\vspace{-2mm}
\paragraph{Realized balanced sample sizes.}
Balanced item counts \emph{per condition} for each language-model pair are shown in Table~\ref{tab:balancedNs}. For each language and model, we downsample all conditions to the smallest valid per-condition count in that language (one masked target per sentence), so $N$ can differ across models because validity is model-dependent: the six-piece span cap and reconstruction filters operate on model-specific subword segmentations (WordPiece vs. SentencePiece/BPE), so the intersection of items that survive \emph{all} conditions can be smaller for one model than the other (e.g., a target that is 7 WordPiece pieces but 5 SentencePiece pieces). Variation across seeds was negligible. Turkish yields markedly smaller $N$ because multi-piece targets and strict word-level reconstruction make many items ineligible across conditions, shrinking the intersection needed for balancing. 

\begin{table}[t]
\centering
\footnotesize
\setlength{\tabcolsep}{4pt}
\renewcommand{\arraystretch}{1.0}
\begin{tabular}{lcc}
\toprule
Lang & $N$ (mBERT) & $N$ (XLM-R) \\
\midrule
DE & 300 & 300 \\
EN & 290 & 304 \\
ES & 300 & 304 \\
RU & 286 & 304 \\
ZH & 300 & 304 \\
\bottomrule
\end{tabular}
\caption{Balanced per-condition counts ($N$) for five languages.}
\label{tab:balancedNs}
\end{table}

\vspace{-2mm}
\paragraph{Tokenization and counting.}
All counting is sentence-based (one masked word per sentence). We report accuracy at the word level by masking and
predicting \emph{entire} word spans (combining all subword pieces; see Sec.~\ref{sec:methodology}), ensuring
comparability between WordPiece (mBERT) and SentencePiece/BPE (XLM-R). 

\subsection{Reproducibility and Code Availability}

Sampling and permutations are controlled by fixed seeds. For a given seed, the same sentence IDs, masked targets, and perturbation permutations are used across models and conditions. We release the code, sentence lists, condition assignments, and per-run JSONL outputs with the paper.\footnote{\url{https://github.com/bondfeld/WordPrediction\_LREC2026}} Target selection does not pre-filter by span length; the cap is enforced at scoring time. The punctuation sets and closed-class stoplists used as safety filters are released with the code.


\subsection{Pretrained Models and Tokenization}

\vspace{-2mm}
\paragraph{mBERT} The multilingual BERT \citep{devlin-etal-2019-bert} is a transformer-based MLM pretrained on Wikipedia corpora from 104 languages. It has 12 transformer layers, 768 hidden units, and 12 attention heads (110M parameters total). Tokenization is performed with a shared WordPiece vocabulary of 119{,}547 subword units across languages. The MLM objective masks 15\% of input tokens, with the model trained to recover the original tokens from context. We use the original, unfine-tuned model. Note: pretraining coverage is uneven across languages (Wikipedia size varies), so cross-language baselines can reflect both modeling and data exposure; we therefore interpret such differences cautiously (see Sec. \ref{sec:limitations}).

\vspace{-2mm}
\paragraph{XLM-R} XLM-RoBERTa base \citep{conneau-etal-2020-unsupervised} is a transformer-based MLM pretrained on CommonCrawl-based CC-100 corpora from 100 languages, with substantially larger pretraining data than mBERT. The base architecture also has 12 layers, 768 hidden units, and 12 attention heads (270M parameters total), but uses a shared SentencePiece/BPE vocabulary of 250{,}002 subword units. Like mBERT, it is trained with a 15\% masking rate, but uses the \texttt{<mask>} token rather than \texttt{[MASK]}. As with mBERT, CC-100 coverage is imbalanced across languages; tokenization and pretraining exposure can therefore contribute to cross-language differences (see Sec. \ref{sec:limitations}).

\vspace{-2mm}
\paragraph{Masking and scoring.}
Masking is performed at the \emph{word level} to ensure comparability between WordPiece (mBERT) and SentencePiece/BPE (XLM-R). The target word is identified in the untokenized sentence, and all of its subword pieces are replaced by the model’s mask token (\texttt{[MASK]} or \texttt{<mask>}). Perturbations are applied to the masked input, and each sentence contains exactly one masked content word. Predictions are reconstructed to full words from subword outputs with a cap of \emph{six} pieces; items exceeding the cap are excluded from word-level evaluation. A prediction is counted correct only if the reconstructed form exactly matches the gold target \emph{after Unicode NFKC normalization, case-folding for alphabetic scripts, and punctuation stripping}. The gold target is the \emph{surface form} for non-\texttt{+L} conditions and the \emph{lemma} for \texttt{+L}.

\section{Baseline}\label{sec:baseline}

Our baseline is the original, unfine-tuned pretrained model evaluated on unmodified sentences (Orig condition) with a single masked target word. For each sentence in the test set, we select one content word from the untokenized text and replace \emph{all} of its subword pieces with the model’s mask token (\texttt{[MASK]} for mBERT, \texttt{<mask>} for XLM-R). 
This word-level masking is deterministic within a seed: the same \emph{target word} is used across models and conditions; under scrambling its position changes with the permutation. Across seeds, target selection may differ.

Predictions are generated with the model’s masked language modeling head. For multi-piece targets, all predicted subwords must match in order for the reconstruction to be marked correct. Accuracy is computed at the \emph{word level} (exact match of reconstructed form after Unicode normalization and case-folding for alphabetic scripts). This fixed-mask baseline serves as the reference point for measuring degradation under the perturbation conditions described in Sec.~\ref{sec:methodology}.

\section{Results}\label{sec:results}

\begin{figure}[t]
  \centering
  \includegraphics[width=0.78\linewidth]{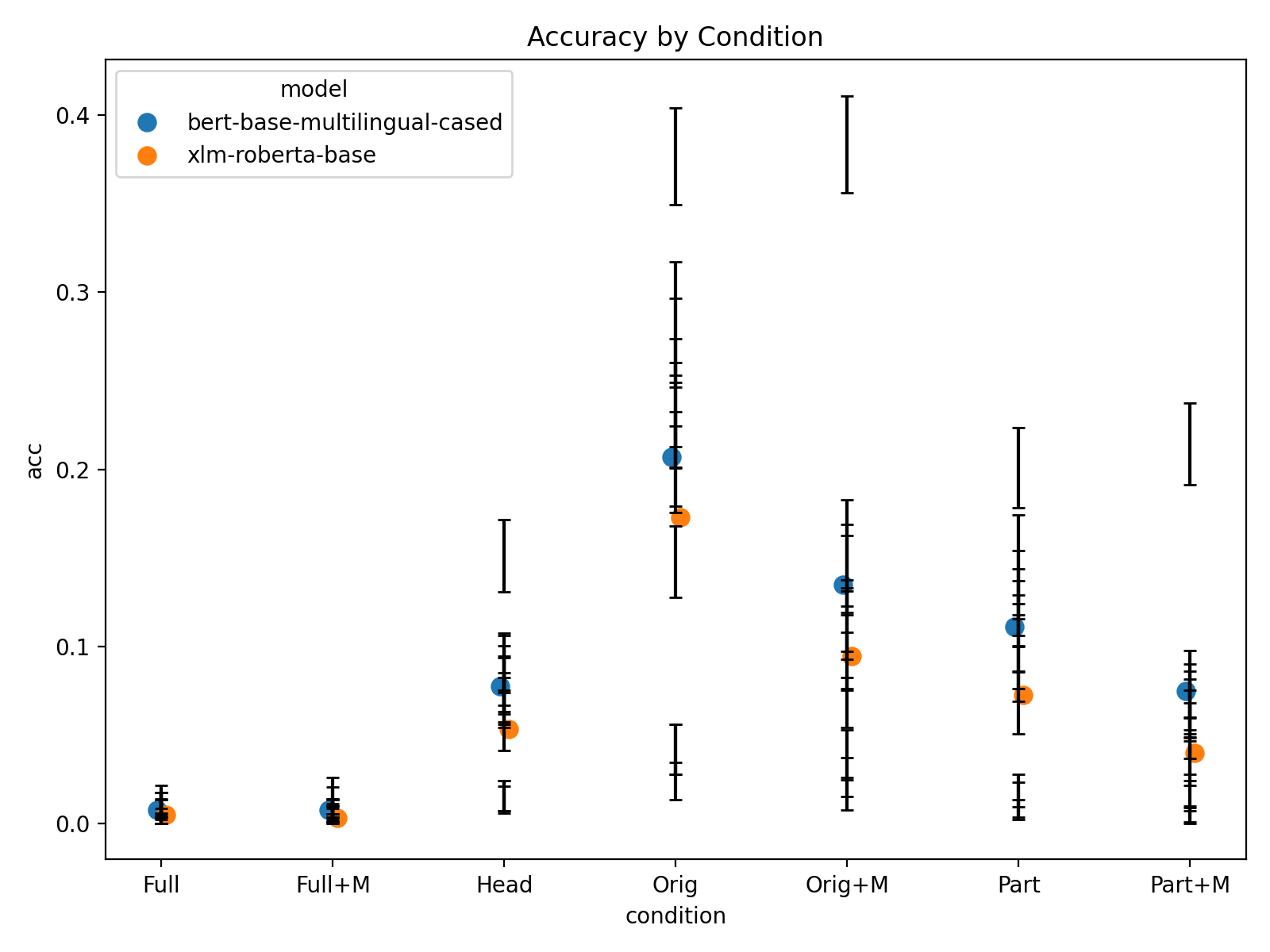}
  \caption{Accuracy across perturbation conditions with Wilson 95\% CIs.}
  \label{fig:accuracy}
\end{figure}

\begin{figure}[t]
  \centering
  \includegraphics[width=0.78\linewidth]{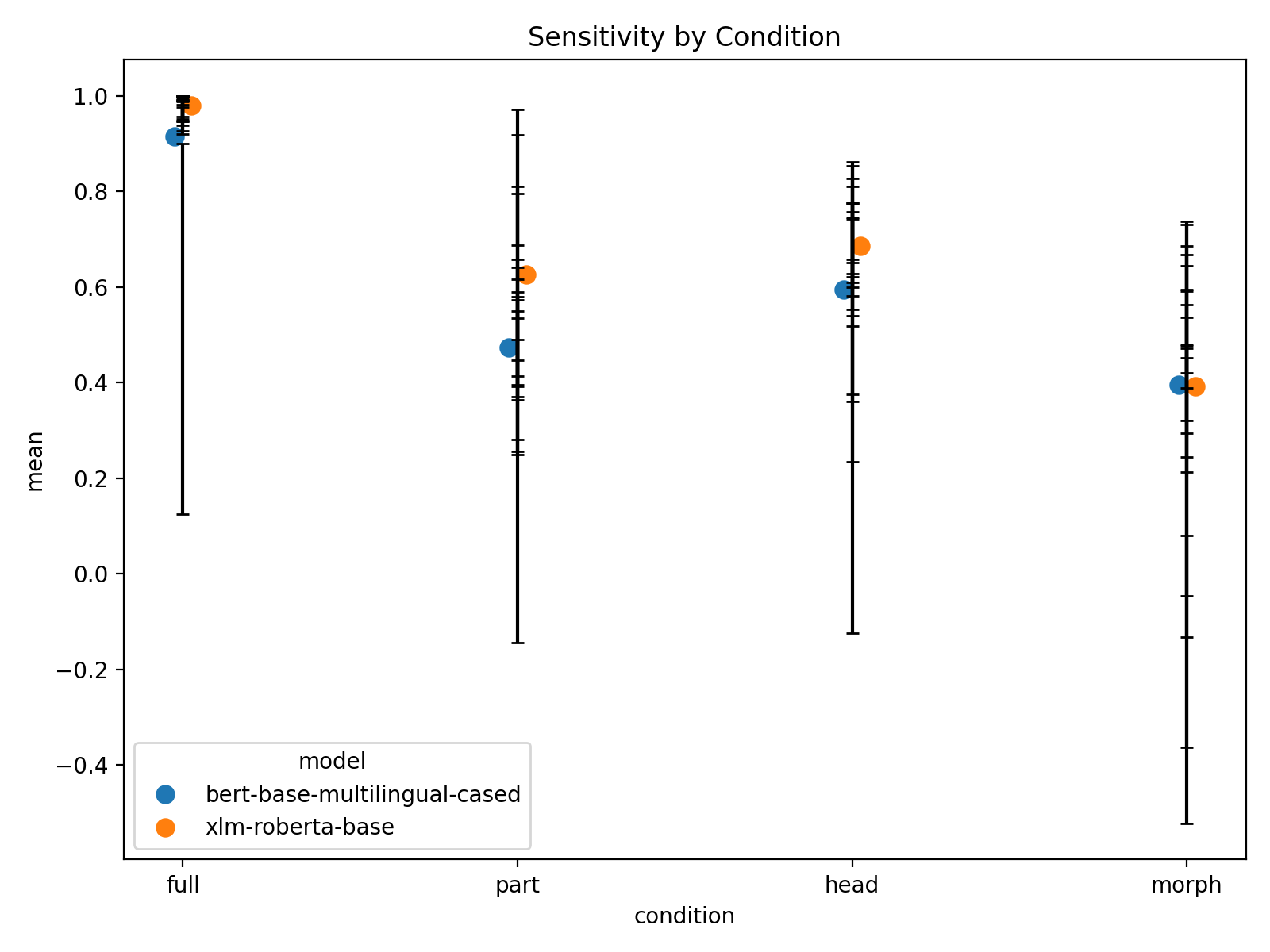}
  \caption{Relative-drop sensitivities $S$ (95\% CIs).}
  \label{fig:sensitivity}
\end{figure}

\begin{figure}[t]
  \centering
  \includegraphics[width=0.78\linewidth]{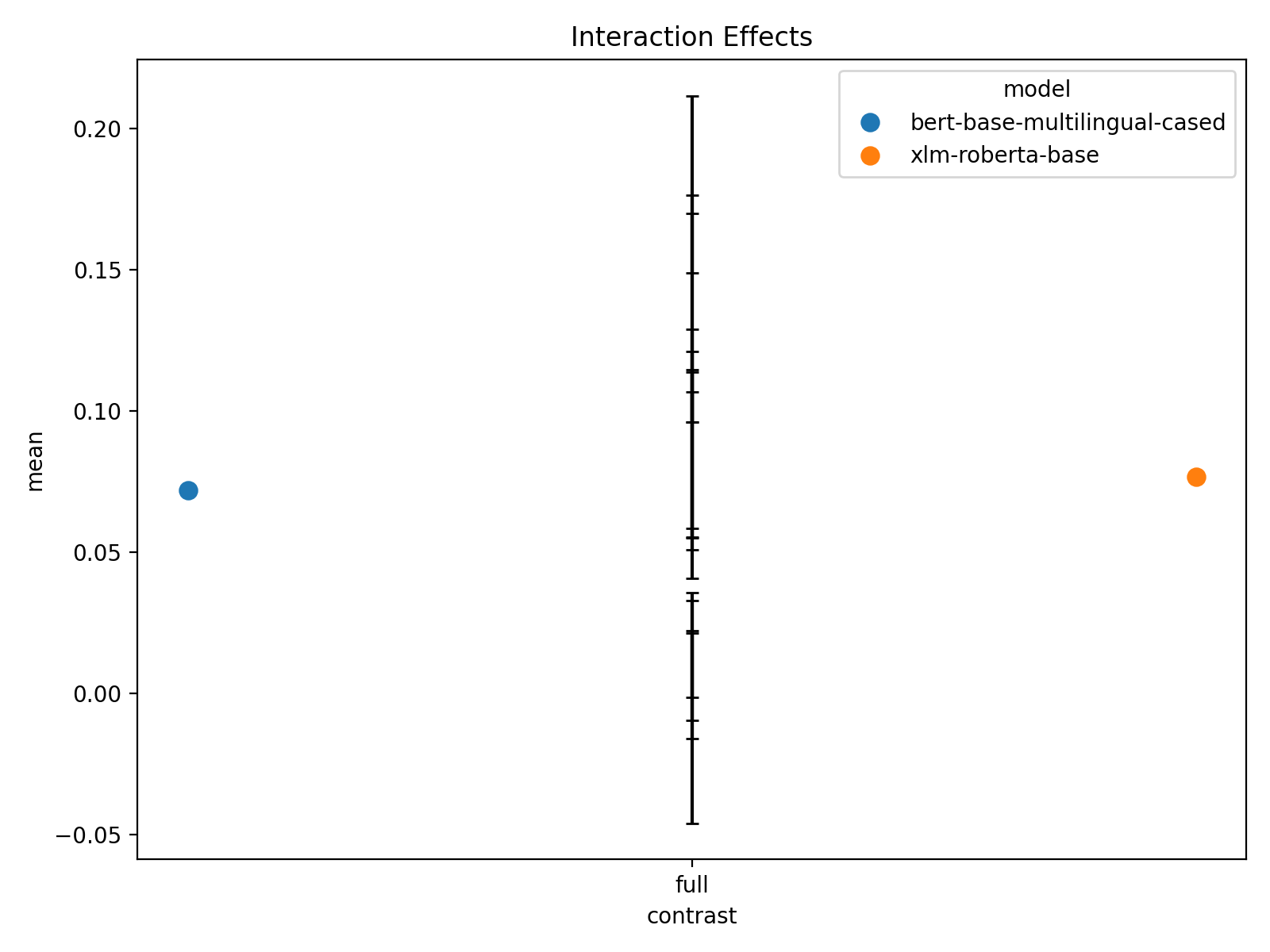}
  \caption{Interaction effects $I_{\text{full}}$ (95\% CIs).}
  \label{fig:interaction}
\end{figure}

We report balanced word-level accuracy by language and condition for mBERT and XLM-R. Main-text tables include DE/EN/ES/RU/ZH; Turkish (TR) is reported in Appendix A due to floor effects under strict reconstruction. Conditions: Orig (original), Full (full permutation), Part (subset of content words permuted), Head (each UD head is swapped with one of its dependents in a single pass (seed-deterministic); other tokens remain in place); +L denotes \emph{sentence-level} UD lemma substitution. Values are averaged over seeds on the balanced sets. We report 95\% confidence intervals: Wilson CIs for accuracy and parametric bootstrap (2,000 draws) for sensitivities $S$ and interaction $I$ (see Figs.~\ref{fig:accuracy}, \ref{fig:sensitivity}, \ref{fig:interaction}).

In addition to top-1 exact reconstruction, we report top-5 word accuracy to capture whether the gold target remains among high-ranked candidates under perturbation (Appendix B).
Top-5 results preserve the main conclusion: full scrambling collapses performance even when allowing multiple candidates.

\begin{table}[t]
\centering
\footnotesize
\setlength{\tabcolsep}{2pt}
\renewcommand{\arraystretch}{1.0}
\resizebox{\columnwidth}{!}{%
\begin{tabular}{lccccccc}
\toprule
\multicolumn{8}{c}{\textbf{mBERT}}\\
\midrule
Lang & Orig & Full & Part & Head & Orig+L & Full+L & Part+L \\
\midrule
DE & 0.143 & 0.013 & 0.075 & 0.078 & 0.060 & 0.003 & 0.033 \\
EN & 0.219 & 0.016 & 0.123 & 0.060 & 0.167 & 0.019 & 0.090 \\
ES & 0.251 & 0.013 & 0.143 & 0.063 & 0.143 & 0.003 & 0.085 \\
RU & 0.276 & 0.008 & 0.119 & 0.063 & 0.119 & 0.008 & 0.061 \\
ZH & 0.443 & 0.023 & 0.223 & 0.133 & 0.448 & 0.023 & 0.230 \\
\bottomrule
\end{tabular}}
\caption{Balanced word-level accuracy for mBERT. “+L” = sentence-level UD lemma substitution (context and label).}
\label{tab:acc_mb}
\end{table}

\begin{table}[t]
\centering
\footnotesize
\setlength{\tabcolsep}{2pt}
\renewcommand{\arraystretch}{1.0}
\resizebox{\columnwidth}{!}{%
\begin{tabular}{lccccccc}
\toprule
\multicolumn{8}{c}{\textbf{XLM-R}}\\
\midrule
Lang & Orig & Full & Part & Head & Orig+L & Full+L & Part+L \\
\midrule
DE & 0.338 & 0.013 & 0.195 & 0.168 & 0.150 & 0.008 & 0.078 \\
EN & 0.303 & 0.006 & 0.166 & 0.079 & 0.218 & 0.003 & 0.123 \\
ES & 0.338 & 0.013 & 0.175 & 0.100 & 0.181 & 0.010 & 0.095 \\
RU & 0.293 & 0.013 & 0.147 & 0.078 & 0.126 & 0.015 & 0.068 \\
ZH & 0.325 & 0.025 & 0.150 & 0.110 & 0.333 & 0.025 & 0.150 \\
\bottomrule
\end{tabular}}
\caption{Balanced word-level accuracy for XLM-R. “+L” = sentence-level UD lemma substitution (context and label).}
\label{tab:acc_xlm}
\end{table}

\noindent To quantify perturbation impact, we use sensitivity to measure the relative loss from \textit{Orig}:
\vspace{-1mm}
{\setlength{\abovedisplayskip}{4pt}\setlength{\belowdisplayskip}{4pt}%
\[
S_{\text{cond}}=\frac{A_{\text{Orig}}-A_{\text{cond}}}{A_{\text{Orig}}}
\]}
\noindent and interaction to measure deviation from additivity for \textit{cond}+L:
\vspace{-1mm}
{\setlength{\abovedisplayskip}{4pt}\setlength{\belowdisplayskip}{4pt}%
\[
I_{\text{cond}}=A_{\text{cond+L}}-\bigl(A_{\text{cond}}+A_{\text{Orig+L}}-A_{\text{Orig}}\bigr).
\]}

Note that \(I\) is computed in accuracy space and depends on the label definition. Because \(I\) mixes \emph{surface-label} terms (Orig, Full) with \emph{lemma-label} terms (Orig+L, Full+L), it is sensitive to that choice. We report \(I\) on the balanced sets in Tables~\ref{tab:sens_mb}-\ref{tab:sens_xlm} and include unbalanced counterparts in the appendix.

\vspace{4pt}
\noindent\textbf{Interpreting $I$ (overlap vs.  synergy).}
Let $D_X\!=\!A_{\text{Orig}}-A_X$ denote harm (in particular $D_{\text{+L}}=A_{\text{Orig}}-A_{\text{Orig+L}}$). Then $I_{\text{full}}=D_{\text{Full}}+D_{\text{+L}}-D_{\text{Full+L}}$. That is, the interaction measure computes the difference between combining the two types of manipulations to using each of them individually. Thus $I>0$ indicates \emph{sub-additive} harm in accuracy space; outside floor/ceiling regimes this pattern is compatible with overlapping/partially redundant cues, whereas $I<0$ indicates \emph{supra-additive} harm (synergy/complementarity).

\vspace{-2mm}
\paragraph{Why these metrics.}
Relative-drop sensitivities $S$ factor out language-model baseline differences $A_{\text{Orig}}$, enabling comparisons across typologically diverse settings. We compute the interaction $I$ in \emph{accuracy} space to test (non-)additivity: $I<0$ means the combined perturbation (Full+L) harms \emph{more} than the additive expectation (supra-additive harm); $I\approx 0$ is additive; $I>0$ indicates \emph{sub-additive} effects in accuracy space. Because both $S$ and $I$ depend on $A_{\text{Orig}}$ and are bounded by floor/ceiling effects, floor cases (e.g., Turkish) compress dynamic range; accordingly, we also report unbalanced results in the appendix and avoid over-interpreting $I$ when baselines are near zero.

\begin{table}[t]
\centering
\footnotesize
\setlength{\tabcolsep}{3pt}
\renewcommand{\arraystretch}{1.0}
\resizebox{\columnwidth}{!}{%
\begin{tabular}{lccccc}
\toprule
\multicolumn{6}{c}{\textbf{mBERT}}\\
\midrule
Lang & $S_{\text{full}}$ & $S_{\text{part}}$ & $S_{\text{head}}$ & $S_{\text{+L}}$ & $I_{\text{full}}$ \\
\midrule
DE & 0.912 & 0.474 & 0.456 & 0.579 & 0.073 \\
EN & 0.925 & 0.438 & 0.725 & 0.238 & 0.055 \\
ES & 0.950 & 0.430 & 0.750 & 0.430 & 0.098 \\
RU & 0.972 & 0.569 & 0.771 & 0.569 & 0.157 \\
ZH & 0.949 & 0.497 & 0.701 & $-0.011$ & $-0.005$ \\
\bottomrule
\end{tabular}}
\caption{Balanced sensitivities and interaction for mBERT computed in \emph{accuracy} space.}

\label{tab:sens_mb}
\end{table}

\begin{table}[t]
\centering
\footnotesize
\setlength{\tabcolsep}{3pt}
\renewcommand{\arraystretch}{1.0}
\resizebox{\columnwidth}{!}{%
\begin{tabular}{lccccc}
\toprule
\multicolumn{6}{c}{\textbf{XLM-R}}\\
\midrule
Lang & $S_{\text{full}}$ & $S_{\text{part}}$ & $S_{\text{head}}$ & $S_{\text{+L}}$ & $I_{\text{full}}$ \\
\midrule
DE & 0.963 & 0.422 & 0.504 & 0.556 & 0.183 \\
EN & 0.982 & 0.450 & 0.739 & 0.279 & 0.082 \\
ES & 0.963 & 0.481 & 0.704 & 0.467 & 0.155 \\
RU & 0.957 & 0.500 & 0.733 & 0.569 & 0.169 \\
ZH & 0.923 & 0.538 & 0.662 & $-0.023$ & $-0.008$ \\
\bottomrule
\end{tabular}}
\caption{Balanced sensitivities and interaction for XLM-R in \emph{accuracy} space.}
\label{tab:sens_xlm}
\end{table}

\begin{figure}[t]
  \centering
  \includegraphics[width=0.78\linewidth]{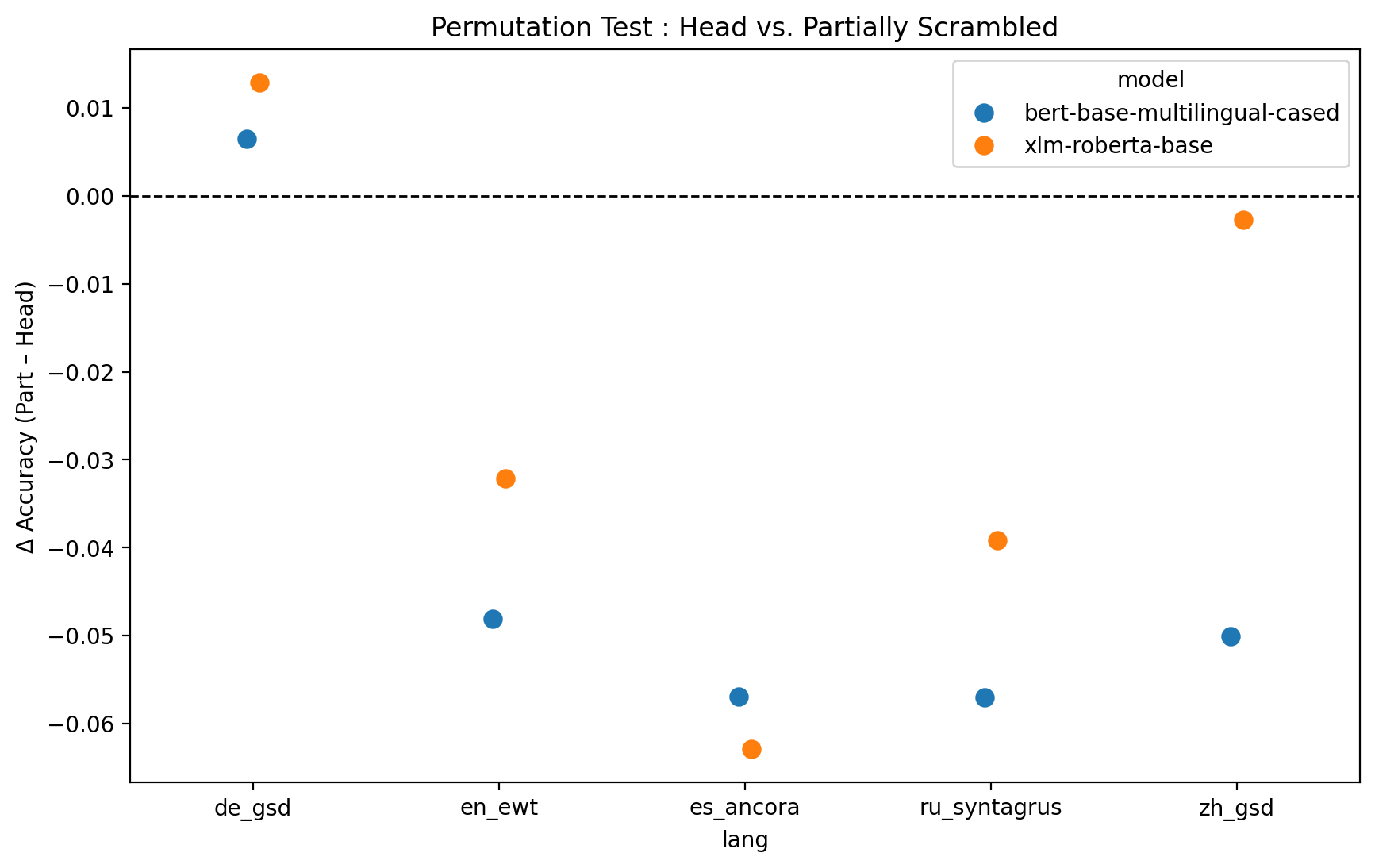}
 \caption{Head vs. Part ($\Delta$ accuracy = Part\,$-$\,Head).}

  \label{fig:head-vs-part}
\end{figure}

\section{Findings and Discussion}\label{sec:discussion}

Because our scrambling manipulations inevitably conflate loss of linear order with reduced well-formedness, observed effects reflect both factors. We report balanced counts and note floor cases in the appendix; isolating “well-formed-only” perturbations is left for future work.

\vspace{2pt}
\noindent\textit{Word order dominates.}
Across balanced sets, \emph{full} scrambling drives word-level accuracy close to zero for both models in every language (Tables~\ref{tab:acc_mb}-\ref{tab:acc_xlm}; Fig.~\ref{fig:accuracy}). \emph{Partial} and \emph{head} scrambling also yield large losses (Fig.~\ref{fig:sensitivity}). This conclusion holds under top-5 evaluation as well: even allowing five candidates, Full scrambling yields very low accuracy across languages (at most 0.053 for mBERT and 0.067 for XLM-R on the balanced sets), indicating that perturbations often push the gold target out of the top candidate set rather than only lowering its rank.

English and Chinese show especially large losses under structured disruption (notably \textit{Head}); German is comparatively less affected in several structured conditions, while Russian remains highly order-sensitive under \textit{Full} despite rich case marking.
Within the structured perturbations, \textit{Head} tends to hurt English more than \textit{Part}, while German is comparatively less affected, consistent with available morphological and function-word cues. We interpret claims conservatively in floor regimes (e.g., strict span reconstruction in Turkish).

\vspace{2pt}
\noindent
\emph{Scramble $+$ Lemma (+L) is sub-additive in accuracy space (with a floor caveat).}
In DE/EN/ES/RU, $I_{\text{full}}$ is positive (Tables~\ref{tab:sens_mb}-\ref{tab:sens_xlm}), i.e., accuracy under \textit{Full+L} is higher than the additive baseline $A_{\text{Full}} + A_{\text{Orig+L}} - A_{\text{Orig}}$. However, because \textit{Full} drives accuracy close to zero, the additive baseline is often negative, and $I_{\text{full}}$ is mechanically biased upward by the 0-floor; we therefore interpret $I_{\text{full}}$ cautiously as a coarse non-additivity diagnostic rather than a clean measure of cue overlap. Consistent with this, on the unbalanced sets the corresponding \textit{Part} interactions are also positive in DE/EN/ES/RU (Appendix Table~\ref{tab:sens_unbal_both}), where floor effects are less severe. In ZH, +L is nearly identity (Table~\ref{tab:perturb_magnitude}), and $I_{\text{full}}$ is near zero.


\vspace{2pt}
\noindent
\emph{Heads vs. partial content shuffles.}
Head–dependent disruption affects languages asymmetrically. English shows a larger Head vs. Part gap than German (consistent with fixed SVO and light inflection). In most other languages (ES/RU/ZH), \textit{Head} is also more harmful than \textit{Part} for both models (Fig.~\ref{fig:head-vs-part}). German exhibits a small exception for mBERT (Head $\approx$ Part), while for XLM-R, \textit{Head} is slightly more harmful than \textit{Part}.

\vspace{2pt}
\noindent
\emph{Why can $I_{\text{full}}$ be positive?}
Under \textit{Full} scrambling, $A_{\text{Full}}$ is often near 0, so the additive baseline
$A_{\text{Full}} + A_{\text{Orig+L}} - A_{\text{Orig}}$ can fall below 0; because accuracy is bounded below by 0, $I_{\text{full}}$ is mechanically biased upward in these floor regimes.
We therefore treat $I_{\text{full}}$ as a coarse non-additivity diagnostic and corroborate it with less floor-bound interactions (e.g., $I_{\text{part}}$ on unbalanced sets; Appendix Table~\ref{tab:sens_unbal_both}).
Outside strong floor/ceiling regimes, positive $I$ is compatible with partially overlapping evidence from word order and inflectional form, but we avoid reading $I_{\text{full}}>0$ as a clean cue decomposition in the near-collapse setting.

\emph{Why is $I_{\text{full}}>0$?}

\vspace{2pt}
\noindent
\emph{Why does +L help (slightly) in ZH for mBERT?}
In Chinese, lemma substitution is vacuous because surface forms $\approx$ lemmas. The tiny uptick (Orig $0.443{\to}0.448$; Full unchanged at $0.023$) reflects tokenization/normalization quirks in mBERT, not genuine morphology.

\vspace{2pt}
\noindent
\emph{Lemma substitution (+L, sentence-level).}
In +L we replace \emph{every token} with its UD \texttt{LEMMA} (CoNLL-U; identity fallback), and evaluate against the lemma of the masked target. Thus, +L lemmatizes both context and target.



\vspace{2pt}
\noindent
\emph{Model-specific notes.} (1) Chinese. On the balanced sets, both models achieve substantive baselines (mBERT~0.443, XLM-R~0.325; Tables~\ref{tab:acc_mb}-\ref{tab:acc_xlm}). +L has negligible effect overall, and $I_{\text{full}}$ is near zero or slightly negative (Tables~\ref{tab:sens_mb}-\ref{tab:sens_xlm}), suggesting limited supra-additivity between order and lemma substitution under our reconstruction protocol. (2) Turkish. Both models are near floor even on Orig; every sensitivity/interaction statistic on TR should be read with floor effects in mind.

\vspace{3pt}
\noindent
\emph{What this means.}
These multilingual MLMs  are still positional workhorses. Morphology on the target helps a bit (sometimes more in DE/ES/RU), but it does \emph{not} compensate for lost order. Head ordering of core arguments matters a lot in English; German’s morphology together with function words makes head swaps slightly less harmful. Cross-model differences (e.g., ZH) are large enough that claims about “the model” need to specify which one.

\noindent \textbf{Lessons learned.} Two themes stand out. First, non-additivity: in DE/ES/RU, we observe $I_{\text{full}}>0$ for both models (e.g., XLM-R: $0.183/0.155/0.169$; mBERT: $0.073/0.098/0.157$), meaning the loss under \textit{Full+L} is smaller than the sum of the separate losses from \textit{Full} and \textit{+L}. This is consistent with overlapping/partially redundant signals from word order and the target’s morphology rather than complementary amplification. Second, head-dependent disruption is asymmetric across languages: English is hit especially hard (fixed SVO, light inflection), whereas German shows mild cushioning. In ES/RU/ZH, \textit{Head} typically hurts more than \textit{Part} for both models; in German this holds for XLM-R (Part $0.195$ vs.\ Head $0.168$) but not for mBERT (Part $0.075$ vs.\ Head $0.078$, i.e., roughly equal). In Chinese, unlike the other languages, \textit{+L} does not reduce the baseline (mBERT: $0.443\!\rightarrow\!0.448$; XLM-R: $0.325\!\rightarrow\!0.333$); we attribute these tiny upticks to tokenization/reconstruction effects rather than genuine morphological leverage, and \textit{+L} never compensates for the loss of order. Finally, Turkish is dominated by floor effects: very low \textit{Orig} baselines and balancing-induced reductions in $N$ compress dynamic range, rendering $S$ and $I$ numerically unstable; we therefore defer to unbalanced summaries in Appendix A.

\section{Conclusion}
Across five languages and two multilingual MLMs (mBERT, XLM-R), word order is the dominant cue. Full scrambling causes near-total collapse in masked word prediction (relative drops $\approx 0.91$-$0.98$ across languages/models); partial and head-only scrambling also inflict large losses. Keeping function words in place does not rescue predictions, and shuffling the relative order of core heads is particularly harmful in English, while German is slightly cushioned by function-word/morphological scaffolding. 

\textit{+L} removes information and generally reduces accuracy; it never compensates for lost order. Any apparent improvements (e.g., tiny upticks for mBERT-ZH) are evaluation artifacts from tokenization/reconstruction, not genuine gains.  Interactions computed in accuracy space are mostly positive in DE/ES/RU ($I_{\text{full}}>0$), indicating sub-additivity (joint harm $<$ additive baseline). Because \textit{Full} accuracy is often near floor, $I_{\text{full}}$ is partly shaped by saturation; we therefore treat it as a coarse non-additivity signal and corroborate it with less floor-bound interactions (e.g., $I_{\text{part}}$ on unbalanced sets; Appendix Table~\ref{tab:sens_unbal_both}).

Baseline behavior matters. mBERT and XLM-R diverge sharply on some languages (notably Chinese, where mBERT’s baseline is much higher), and Turkish baselines are so low that ``drops'' and interaction scores are floor-limited. The evidence indicates strong behavioral sensitivity to word-order disruption in this diagnostic setting and only limited robustness under lemma-normalized contexts.


Two directions look promising: (i) training objectives that reduce over-reliance on absolute order (e.g., de-scrambling or order-invariant auxiliaries) while preserving syntactic signals, and (ii) morphology-aware prediction heads that expose inflectional structure explicitly rather than relying on subword artifacts. Evaluation-wise, adding constituency-based scrambling and human baselines on the same items would sharpen what cues models vs.  humans actually use when order is unreliable.

\section{Limitations}\label{sec:limitations}

We evaluate \emph{Orig}, \emph{Full}, \emph{Part}, and \emph{Head}, plus \emph{+L} variants where used (\emph{Orig+L}, \emph{Full+L}, \emph{Part+L}); \emph{+L} is \emph{sentence-level} lemma substitution: the whole context is lemmatized and the gold label is the lemma of the masked target. We do not include target-only or context-only normalization, and we do not evaluate Head+L.

Exactly one content word is masked per sentence. Within each seed, the target is fixed across conditions (and across models), but targets can differ across seeds. Accuracy is exact word-level reconstruction from subword predictions, which penalizes near-misses and interacts with subword length/frequency, potentially understating partial knowledge. Tokenization differs by model (WordPiece for mBERT; SentencePiece/BPE for XLM-R) and across languages, so some baseline gaps and asymmetries likely reflect segmentation as well as modeling. We partially address this by reporting top-5 word accuracy; however, we do not report full probability shifts (e.g., NLL/entropy) because we do not store full vocab distributions for each item.
Head-only scrambling uses UD \emph{gold} trees: in one seed-deterministic pass we swap each head token with one of its dependents (any label), leaving other tokens fixed. We do not parse at test time and do not perform phrase/constituency-level reordering.

For each language-model pair we downsample every condition to the smallest per-condition count, which reduces $N$; Turkish is most affected because many targets exceed the six-piece reconstruction cap, producing floor effects that make sensitivities and interactions numerically fragile. We report means on balanced sets and compute sensitivities/interactions in \emph{accuracy} space; given the near-collapse under \emph{Full} scrambling, additional significance tests add little beyond visible effect sizes, so we omit $p$-values.

Finally, this is a diagnostic self-contained masked-word recovery study with unfine-tuned models on UD sentences; it is distribution-shifted relative to pretraining and may not predict downstream, fine-tuned behavior. We evaluate two base multilingual encoders (mBERT, XLM-R); extending to newer multilingual LMs is left for future work. Our protocol (balanced sampling, structured scrambling, accuracy-space interactions) transfers directly.

\section*{Acknowledgments}
This research was supported in part by the National Science Foundation under Grant No.~\textit{2226006}.

\section{Bibliographical References}\label{sec:reference}
\bibliography{custom}
\bibliographystyle{lrec-coling2024-natbib}
\newpage
\section*{Appendix A: Turkish (TR) Results}\label{app:tr}

\paragraph{Scope.}
This appendix reports (i) per-language balanced accuracies and $N$’s, (ii) unbalanced summaries to avoid floor/ceiling compression, and (iii) Turkish (TR) results, which are floor-limited under strict word-level reconstruction.

\paragraph{Conventions.}
Word-level accuracy is exact match after Unicode NFKC and case-folding (alphabetic scripts). Confidence intervals are Wilson 95\% CIs for accuracy and parametric 95\% CIs for sensitivities/interactions (2,000 bootstrap draws). “+L” denotes \emph{sentence-level} UD lemma substitution (all tokens lemmatized; target label is the lemma; identity fallback when missing).

\paragraph{Filtering and balancing.}
Each sentence contains one masked content word; items exceeding the six-piece reconstruction cap are excluded. For “Part,” examples with no movement under the content-word shuffle are dropped. Balanced sets downsample each condition to the smallest per-condition count within a language–model pair; unbalanced tables include all valid items.

\paragraph{Computation notes.}
Sensitivities are $S_{\text{cond}}=(A_{\text{Orig}}-A_{\text{cond}})/A_{\text{Orig}}$. Interaction is
$I_{\text{cond}}=A_{\text{cond+L}}-(A_{\text{cond}}+A_{\text{Orig+L}}-A_{\text{Orig}})$,
computed in accuracy space. Head scrambling swaps each UD head token with one dependent in a single, seed-deterministic pass; other tokens remain fixed.

\paragraph{Background.}
Turkish (TR) exhibits strong floor effects because agglutinative morphology yields long, multi-piece targets that often exceed the six-piece reconstruction cap. After filtering, the remaining items sit near floor. We emphasize unbalanced TR summaries and interpret absolute differences cautiously.

\begin{table}[h]
\centering
\footnotesize
\setlength{\tabcolsep}{5pt}
\renewcommand{\arraystretch}{1.0}
\begin{tabular}{lcc}
\toprule
 & $N$ (mBERT) & $N$ (XLM-R) \\
\midrule
TR (balanced per condition) & 108 & 152 \\
\bottomrule
\end{tabular}
\caption{Balanced per-condition counts ($N$) for Turkish (TR).}
\label{tab:tr_balancedN}
\end{table}

\begin{table}[h]
\centering
\footnotesize
\setlength{\tabcolsep}{2pt}
\renewcommand{\arraystretch}{0.9}
\begin{tabular}{lccccccc}
\toprule
\multicolumn{8}{c}{\textbf{TR: Balanced word-level accuracy}}\\
\midrule
Model & Orig & Full & Part & Head & Orig+L & Full+L & Part+L \\
\midrule
mBERT & 0.000 & 0.000 & 0.019 & 0.006 & 0.013 & 0.013 & 0.019 \\
XLM-R & 0.063 & 0.000 & 0.057 & 0.013 & 0.006 & 0.006 & 0.019 \\
\bottomrule
\end{tabular}
\caption{Balanced word-level accuracies for Turkish (TR). Interpret with caution (floor, selection, reconstruction effects).}
\label{tab:tr_bal_acc}
\end{table}

\noindent
\textit{Note on Table~\ref{tab:tr_bal_acc}.} In TR we sometimes see $A_{\text{cond}}>A_{\text{Orig}}$; with near-floor baselines and balanced downsampling (span-cap and “no-move” filters), these tiny deltas are noise rather than real gains from scrambling.

\begin{table}[h]
\centering
\footnotesize
\setlength{\tabcolsep}{3pt}\renewcommand{\arraystretch}{1.0}
\caption{Sensitivities and interaction in \emph{accuracy} space on \emph{unbalanced} sets (all languages shown for cross-reference).}
\label{tab:sens_unbal_both}
\begin{tabular}{lcccccc}
\toprule
\multicolumn{7}{c}{\textbf{mBERT}}\\
\midrule
\textbf{Lang} & $S_{\text{full}}$ & $S_{\text{part}}$ & $S_{\text{head}}$ & $S_{\text{morph}}$ & $I_{\text{full}}$ & $I_{\text{part}}$ \\
DE & 0.980 & 0.418 & 0.438 & 0.559 & 0.081 & 0.036 \\
EN & 0.954 & 0.457 & 0.706 & 0.323 & 0.073 & 0.026 \\
ES & 0.964 & 0.420 & 0.712 & 0.407 & 0.094 & 0.042 \\
RU & 0.982 & 0.533 & 0.724 & 0.629 & 0.166 & 0.096 \\
TR & 0.719 & 0.450 & 0.465 & 0.143 & 0.004 & 0.000 \\
ZH & 0.958 & 0.490 & 0.655 & -0.005 & 0.000 & 0.005 \\
\midrule
\multicolumn{7}{c}{\textbf{XLM-R}}\\
\midrule
\textbf{Lang} & $S_{\text{full}}$ & $S_{\text{part}}$ & $S_{\text{head}}$ & $S_{\text{morph}}$ & $I_{\text{full}}$ & $I_{\text{part}}$ \\
DE & 0.968 & 0.470 & 0.563 & 0.594 & 0.173 & 0.084 \\
EN & 0.964 & 0.445 & 0.716 & 0.297 & 0.079 & 0.033 \\
ES & 0.977 & 0.498 & 0.715 & 0.458 & 0.148 & 0.072 \\
RU & 0.975 & 0.546 & 0.723 & 0.604 & 0.168 & 0.096 \\
TR & 0.931 & 0.437 & 0.706 & 0.688 & 0.045 & 0.017 \\
ZH & 0.938 & 0.587 & 0.644 & 0.012 & 0.004 & 0.001 \\
\bottomrule
\end{tabular}
\end{table}

\begin{figure}[h]
  \centering
  \includegraphics[width=0.78\linewidth]{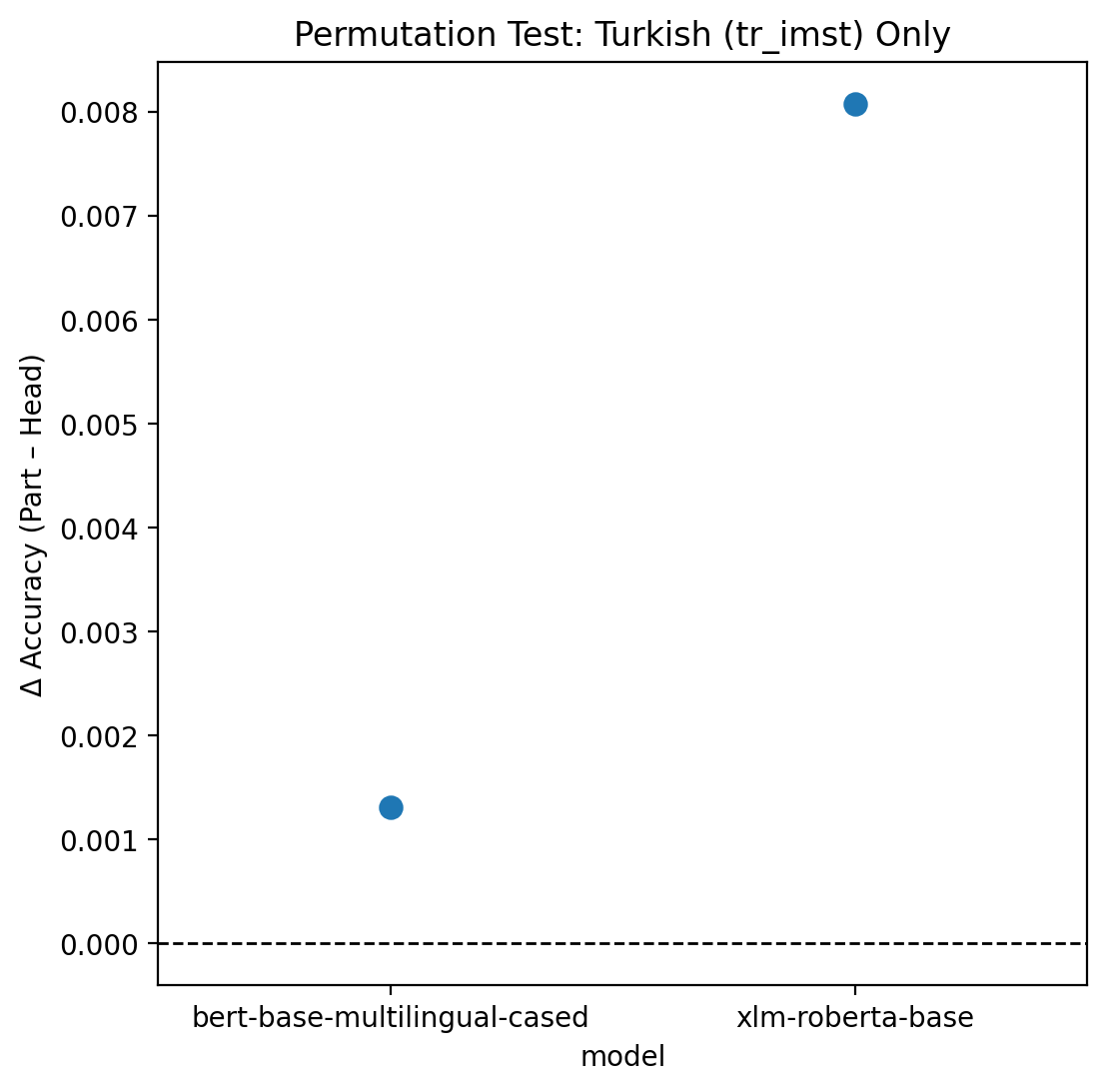}
  \caption{Turkish (TR) only: Head vs.\ Part ($\Delta$ accuracy = Part\,$-$\,Head).}
  \label{fig:tr_head_vs_part}
\end{figure}

\paragraph{Interpretation.}
Both models are at or near floor in TR across all conditions. Because Orig accuracy is close to zero, relative drops and interaction terms are not meaningful indicators of robustness. Micro-differences between conditions are within noise and largely reflect span-cap filtering rather than genuine morphological or order sensitivity.
\newpage
\section*{Appendix B}\label{app:b}

\begin{table}[h]
\centering
\footnotesize
\setlength{\tabcolsep}{2pt}
\renewcommand{\arraystretch}{1.0}
\resizebox{\columnwidth}{!}{%
\begin{tabular}{lccccccc}
\toprule
\multicolumn{8}{c}{\textbf{mBERT (Top-5)}}\\
\midrule
Lang & Orig & Full & Part & Head & Orig+L & Full+L & Part+L \\
\midrule
DE & 0.282 & 0.021 & 0.185 & 0.153 & 0.166 & 0.013 & 0.102 \\
EN & 0.400 & 0.031 & 0.236 & 0.140 & 0.312 & 0.029 & 0.191 \\
ES & 0.423 & 0.033 & 0.264 & 0.170 & 0.266 & 0.037 & 0.175 \\
RU & 0.403 & 0.014 & 0.201 & 0.127 & 0.172 & 0.017 & 0.105 \\
ZH & 0.588 & 0.053 & 0.331 & 0.247 & 0.583 & 0.055 & 0.328 \\
\bottomrule
\end{tabular}}
\caption{Balanced top-5 word accuracy for mBERT.}
\label{tab:top5_mb}
\end{table}

\begin{table}[h]
\centering
\footnotesize
\setlength{\tabcolsep}{2pt}
\renewcommand{\arraystretch}{1.0}
\resizebox{\columnwidth}{!}{%
\begin{tabular}{lccccccc}
\toprule
\multicolumn{8}{c}{\textbf{XLM-R (Top-5)}}\\
\midrule
Lang & Orig & Full & Part & Head & Orig+L & Full+L & Part+L \\
\midrule
DE & 0.503 & 0.030 & 0.311 & 0.289 & 0.281 & 0.018 & 0.171 \\
EN & 0.485 & 0.031 & 0.288 & 0.184 & 0.390 & 0.022 & 0.252 \\
ES & 0.496 & 0.030 & 0.294 & 0.233 & 0.300 & 0.022 & 0.193 \\
RU & 0.439 & 0.022 & 0.262 & 0.184 & 0.224 & 0.025 & 0.153 \\
ZH & 0.464 & 0.067 & 0.263 & 0.237 & 0.466 & 0.066 & 0.263 \\
\bottomrule
\end{tabular}}
\caption{Balanced top-5 word accuracy for XLM-R.}
\label{tab:top5_xlm}
\end{table}

\end{document}